\documentclass[11pt]{article}

\usepackage{acl}
\PassOptionsToPackage{table}{xcolor}
\usepackage{times}
\usepackage{latexsym}
\usepackage[table]{xcolor}
\usepackage{booktabs}
\usepackage{amsmath, amssymb}
\usepackage{soul}

\usepackage{xcolor}
\usepackage{multirow}
\usepackage{enumitem}

\usepackage{booktabs}      
\usepackage{array}         
\usepackage{algorithm}     
\usepackage{algpseudocode} 

\usepackage[T1]{fontenc}
\usepackage[utf8]{inputenc} 
\usepackage{xcolor}
\usepackage[most]{tcolorbox}
\usepackage{fvextra}
\usepackage{subcaption}

\DefineVerbatimEnvironment{PromptText}{Verbatim}{
  fontsize=\footnotesize,
  breaklines=true,
  breakanywhere=false, 
  obeytabs=true,
  tabsize=2,
  baselinestretch=1
}

\newtcolorbox{PromptFigureWide}[2][]{%
  enhanced,
  colback=black!2,
  colframe=black!35,
  boxrule=0.4pt,
  arc=1.6mm,
  left=1.6mm,right=1.6mm,top=1.2mm,bottom=1.2mm,
  boxsep=0.6mm,
  fonttitle=\bfseries\footnotesize,
  title={#2},
  width=\textwidth,
  before skip=0.4\baselineskip,
  after skip=0.4\baselineskip,
  #1
}

\usepackage[most]{tcolorbox}
\tcbuselibrary{listingsutf8} 

\newtcolorbox{mybox3}[1]{colbacktitle=white,coltitle=black,colback=white,colframe=black,fonttitle=\bfseries,fontupper=\small,title=#1,leftupper=0.5em,rightupper=0.5em,boxrule=1.0pt}

\newtcolorbox{PromptBox}[1]{
    colback=gray!10,
    colframe=black!60,
    title=#1,
    fonttitle=\bfseries,
    breakable,
    boxrule=0.5pt,
    arc=2mm,
    fontupper=\footnotesize, 
}

\usepackage[T1]{fontenc}

\usepackage[utf8]{inputenc}

\usepackage{microtype}

\usepackage{inconsolata}

\usepackage{graphicx}

\title{From Answers to Arguments: Toward Trustworthy Clinical Diagnostic Reasoning with Toulmin-Guided Curriculum Goal-Conditioned Learning}

\author{
   ~~Chen Zhan$^{1}$$^{*}$,
   ~~Xiaoyu Tan$^{2}$$^{*}$,
   ~~Gengchen Ma$^{1}$,
   ~~\textbf{Yu-Jie Xiong}$^{1}$, 
   ~~\textbf{Xiaoyan Jiang}$^{1}$,
   ~~\textbf{Xihe Qiu}$^{1}$$^{\dagger}$\\
   $^{1}$School of Electronic and Electrical Engineering, Shanghai University of Engineering Science,\\ Shanghai, 201620, China. \\
   $^{2}$ Tencent Youtu Lab, Shanghai, 200232, China. \\
   \texttt{chenzhan361@gmail.com, arthurtan@tencent.com, qiuxihe1993@gmail.com} \\
   \small{$^{*}$ Equal contribution \hspace{1em} $^{\dagger}$ Corresponding author}
}

\begin{document}
\maketitle

\begin{abstract}
The integration of Large Language Models (LLMs) into clinical decision support is critically obstructed by their opaque and often unreliable reasoning. In the high-stakes domain of healthcare, correct answers alone are insufficient; \textbf{clinical practice demands full transparency to ensure patient safety and enable professional accountability.} A pervasive and dangerous weakness of current LLMs is their tendency to produce\textbf{ "correct answers through flawed reasoning."} This issue is far more than a minor academic flaw; such process errors signal \textbf{a fundamental lack of robust understanding, making the model prone to broader hallucinations and unpredictable failures when faced with real-world clinical complexity.} In this paper, we establish a framework for trustworthy clinical argumentation by adapting the Toulmin model to the diagnostic process. We propose a novel training pipeline: \textbf{Curriculum Goal-Conditioned Learning (CGCL)}, designed to progressively train LLM to generate diagnostic arguments that explicitly follow this Toulmin structure. CGCL's progressive three-stage curriculum systematically builds a solid clinical argument: (1) extracting facts and generating differential diagnoses; (2) justifying a core hypothesis while rebutting alternatives; and (3) synthesizing the analysis into a final, qualified conclusion. We validate CGCL using \textbf{T-Eval}, a quantitative framework measuring the integrity of the diagnosis reasoning. Experiments show that our method achieves diagnostic accuracy and reasoning quality comparable to resource-intensive Reinforcement Learning (RL) methods, while offering a more stable and efficient training pipeline. \footnote{https://github.com/Leonard-zc/CGCL.}

\end{abstract}

\section{Introduction}

\begin{figure}[h]
\centering
  \includegraphics[width=\columnwidth]{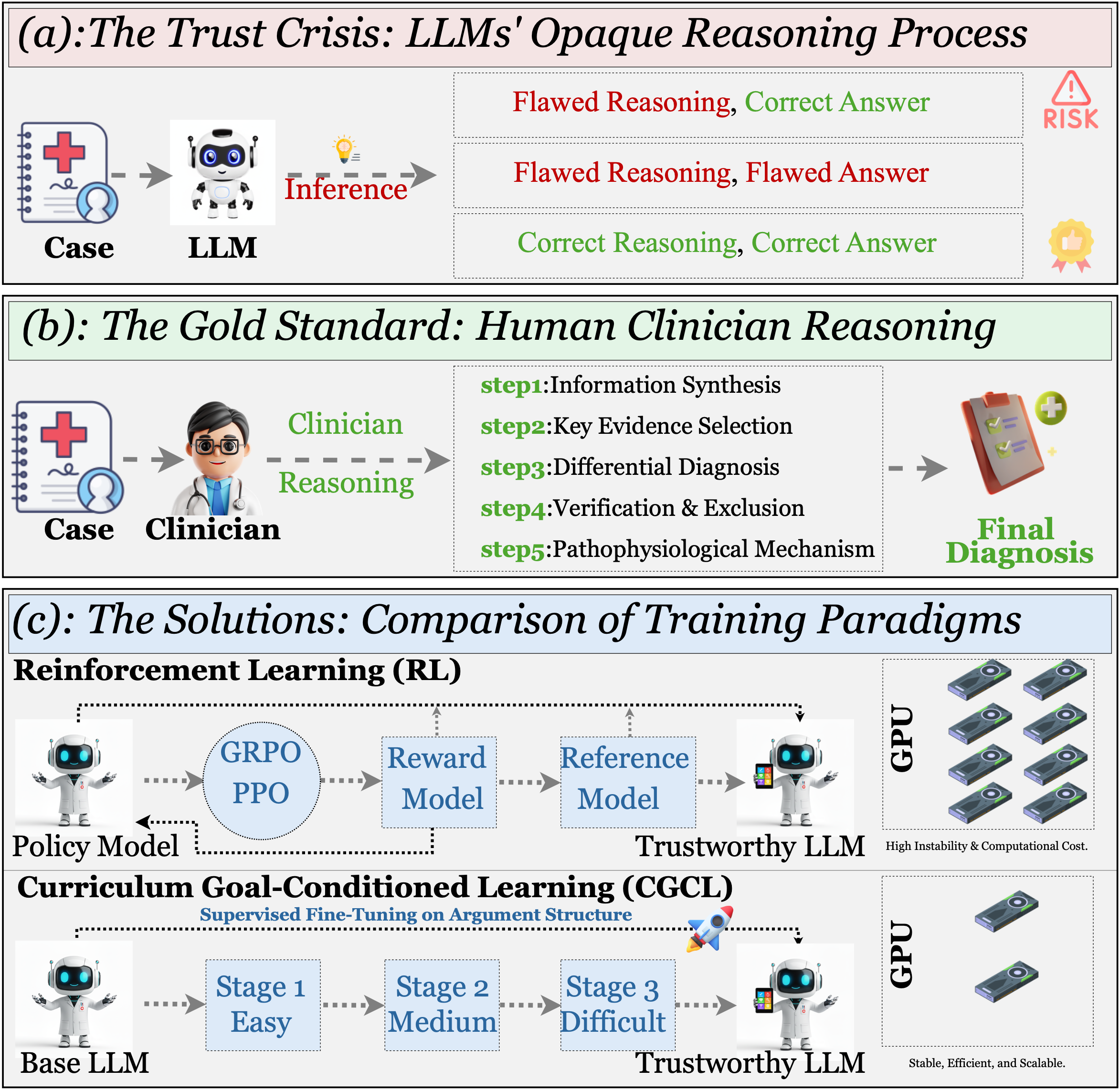}
  \caption{Clinical Diagnostic Reasoning Paradigms.}
  \label{fig1}
\end{figure}

Large Language Models (LLMs) are rapidly reshaping both science and society, and healthcare shows this impact most vividly ~\citep{yu2025large,tan2024chatgpt}. From automating clinical note summarization and assisting in image interpretation to accelerating drug discovery ~\citep{obuchowicz2025artificial,liu2024large}, LLMs are now entering critical stages of the medical pipeline. Recent systems such as GPT-5 and Gemini-2.5-Pro have posted record-breaking results on medical benchmarks like MedQA and the USMLE, in some cases even surpassing human experts \citep{nori2023capabilities,kung2023performance,tu2025towards}. But standardized exams are not the same as real clinical practice. Clinical decision-making is not about picking the right option from a fixed set. Instead, it is about reasoning under uncertainty, integrating messy and incomplete information, and making judgments where the cost of error is human life \citep{hager2024evaluation,mccoy2025assessment}. This leads us to the core scientific question: \textbf{Do today’s LLMs truly have the capacity to function as reliable clinical decision-makers, or are they only test-takers with no license to practice?}

At the heart of this gap is a dangerous illusion: \textbf{models that produce correct answers for the wrong reasons.} Most current evaluation paradigms fixate on whether the final answer is correct, while overlooking the reasoning path that leads there \citep{turpin2023language,lanham2023measuring}. In medicine, however, the path matters as much as the destination. Take a patient with fever and elevated white blood cell count. An LLM might flag a bacterial infection and suggest antibiotics seemingly correct, but only by matching patterns in the text. What it may miss are subtle but critical indicators of sepsis, such as rising lactate levels or borderline hypotension, which demand urgent intervention. In this case, the model’s ``right'' answer is anchored in shallow reasoning that would fail under real clinical pressure \citep{jin2020diseasedoespatienthave}. \textbf{Such brittle correctness is not a minor academic flaw; it is a systemic hazard. By rewarding outcomes without probing reasoning, current benchmarks inflate our perception of what LLMs can actually do.} This misalignment fuels a growing trust crisis and stands as the central obstacle between impressive lab performance and safe, reliable use at the bedside. Figure~\ref{fig1} summarizes this shift from answer-centric evaluation to the structured clinical reasoning paradigm studied in this paper.

\textbf{This trust crisis is rooted not just in model behavior, but in the very way we evaluate and train them.} On the evaluation side, current practices mostly reduce performance to answer accuracy or rely on subjective human ratings \citep{asgari2025framework}. Neither provides a structured, quantitative way to assess the quality of reasoning—the logical steps, evidentiary grounding, and robustness of an argument remain largely invisible. On the training side, mainstream methods such as Supervised Fine-Tuning (SFT) and Reinforcement Learning (RL) inherit the same bias toward final answers \citep{kaufmann2024survey}. The result is a cycle of over-optimizing for correctness while under-optimizing for reasoning. In theory, RL could help address this by shaping richer objectives, but in practice, the barriers are steep: designing reliable reward models is notoriously difficult, training can be unstable, and the computational demands can be substantial in practice. Breaking this bottleneck calls for new paradigms that are both stable and resource-efficient, while directly targeting reasoning as a first-class objective rather than a byproduct \citep{luo2024improve,lightman2023let}.

A natural foundation for clinical reasoning is the \textbf{Toulmin model} of argumentation, which emphasizes how claims must be supported by evidence, qualified by uncertainty, and defended against counter-arguments \citep{toulmin2003uses}. This structure closely mirrors the reasoning clinicians use when moving from symptoms and lab results to a defensible diagnosis \citep{ju2017medical,caroprese2022argumentation}. Yet current LLM pipelines neither evaluate nor learn in this structured way, leaving their ``reasoning'' opaque and often superficial. 

To close this gap, we place the Toulmin model at the core of both evaluation and training. On the evaluation side, we introduce \textbf{T-Eval}, a scalable framework that quantitatively assesses an LLM’s diagnostic reasoning by explicitly measuring the strength of claims, evidence, and rebuttals \citep{yun2025med}. On the training side, we design \textbf{Curriculum Goal-Conditioned Learning (CGCL)}, a pipeline that teaches a model to reason in Toulmin’s structured form, following the progression of clinical training \citep{bengio2009curriculum,schaul2015universal,andrychowicz2017hindsight}. In Stage 1 (Fact Gathering), the model plays the role of a junior resident, extracting findings and proposing an initial differential. In Stage 2 (Hypothesis Testing), it advances to a senior resident, justifying its main hypothesis with physiological evidence while refuting alternatives. In Stage 3 (Synthesis \& Conclusion), it acts as an attending physician, integrating all evidence into a well-qualified final diagnosis. This curriculum can effectively operationalize Toulmin-style reasoning through imitation learning (e.g., SFT) without relying on RL. In experiments on complex real-world medical cases, CGCL-trained models not only achieve diagnostic accuracy competitive with strong baselines, but also substantially improve structured reasoning quality under T-Eval, with the clearest gains on smaller models.

\begin{figure*}[t]
 \centering
  \includegraphics[width=\textwidth]{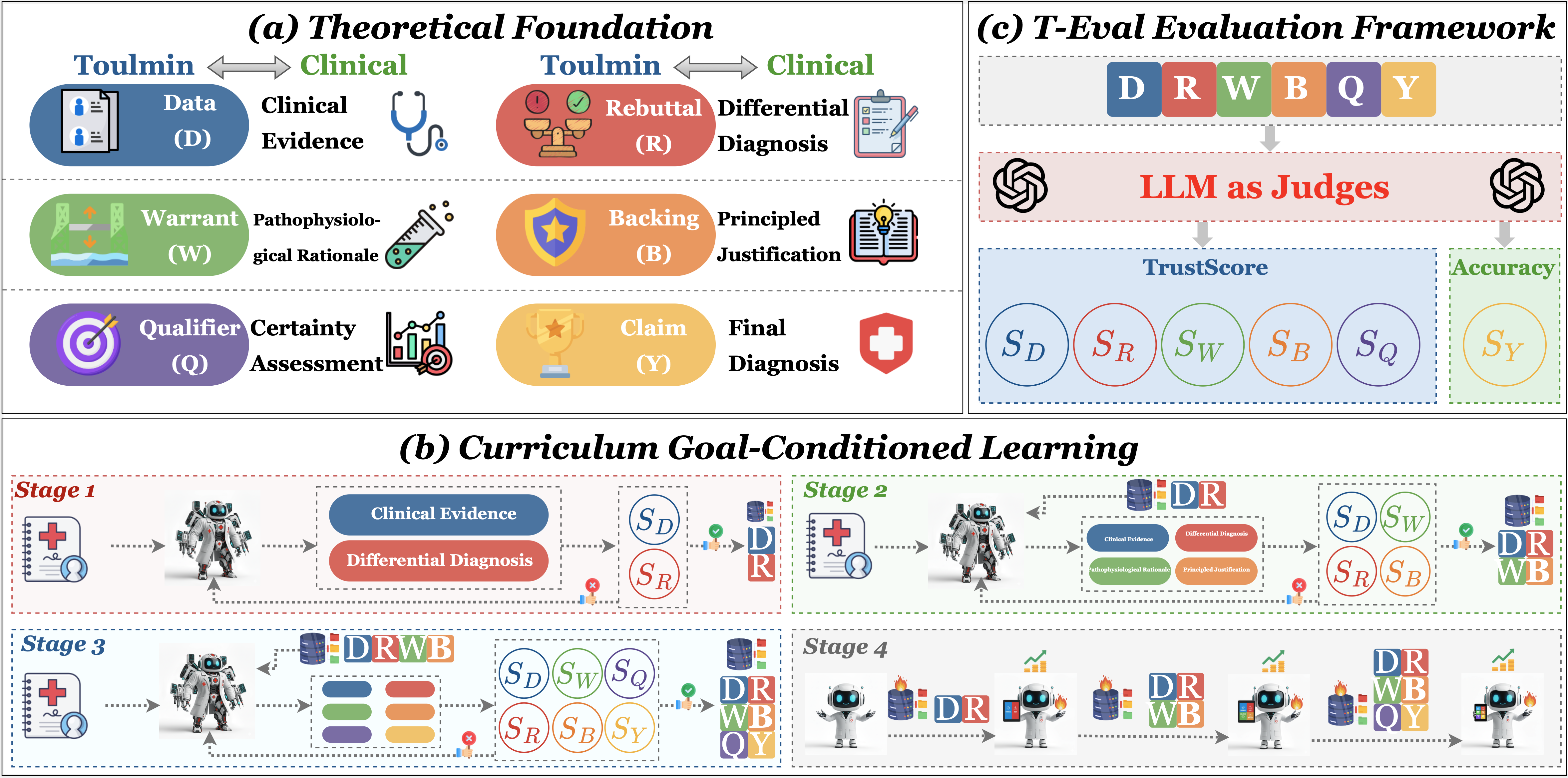}
  \caption{The Synergistic Architecture of CGCL and T-Eval for Trustworthy Clinical Reasoning.}
  \label{fig2}
\end{figure*}

We make the following key contributions:\\
\textbf{(1) T-Eval: A Toulmin-based framework for evaluating clinical reasoning.} We introduce the first scalable method to move beyond answer accuracy and directly measure the structure and integrity of diagnostic arguments, capturing how claims are supported, qualified, and defended.\\
\textbf{(2) CGCL: A training paradigm that instills Toulmin-style reasoning.} We design a three-stage, goal-conditioned curriculum that mirrors medical training, systematically teaching LLMs to gather facts, test hypotheses, and synthesize conclusions in a transparent and stable manner.\\
\textbf{(3) Extensive validation on real-world clinical cases.} Through comprehensive experiments, we demonstrate that a well-structured curriculum can match the reasoning capabilities of complex RL-based approaches while reducing computational overhead.

\section{Methodology}
Our framework bridges the gap between answer accuracy and trustworthy reasoning in clinical LLMs. We teach structured reasoning via CGCL, which builds Toulmin-style diagnostic arguments, and assess it with T-Eval, which scores structural integrity. As illustrated in Figure \ref{fig2}, decomposed sub-tasks yield transparent, goal-guided reasoning.

\subsection{Problem Formulation and Toulmin Model Instantiation}
\label{Toulmin Model}
\subsubsection{Toulmin-structured diagnostic argument.}
We represent clinical diagnostic reasoning as a structured argument
$A=\{D,R,W,B,Q,Y\}$ under the Toulmin model (see Appendix~\ref{app:toulmin}).
Here, $D$ are case-grounded evidence items, $R$ is a ranked differential with brief rebuttals,
$W$ links evidence to hypotheses via pathophysiology, $B$ states supporting clinical principles,
$Q$ calibrates uncertainty and missing information, and $Y$ is the final diagnosis.

\subsubsection{Problem Formulation}
We formalize trustworthy clinical diagnostic reasoning as a \textbf{structured text generation} task. Given a patient case presentation $P = \{p_1, p_2, \ldots, p_m\}$, the objective is to train a model $M_\theta$ to generate a diagnostic argument $A$ that explicitly instantiates the Toulmin structure through a sequence of intermediate outputs $C^{(k)}$
\begin{equation}
A = f(P) = \{D, R, W, B, Q, Y\}
\end{equation}
where the argument is constructed progressively through three curriculum stages:
\begin{equation}
\begin{aligned}[t]
C^{(1)} &= \{D, R\} \\
C^{(2)} &= C^{(1)} \cup \{W, B\} \\
C^{(3)} &= C^{(2)} \cup \{Q, Y\}
\end{aligned}
\end{equation}
The quality of argument $A$ is quantified by our T-Eval framework, which assesses the integrity of all Toulmin components. The learning objective is to find optimal parameters that maximize the expected argument quality:
\begin{equation}
\theta^* = \arg \max_{\theta} \, \mathbb{E}_{P \sim \mathcal{P}} \left[ T\text{-}Eval(A) \mid A = M_\theta(P) \right]
\end{equation}
This formulation directly motivates our curriculum learning approach in Section~\ref{sec:cgcl}, where we sequentially optimize for the generation of each component set $C^{(k)}$.

\subsection{Curriculum Goal-Conditioned Learning}
\label{sec:cgcl}
The CGCL framework is architected around the principle of \textbf{Goal-Conditioned Offline Imitation Learning}. This paradigm sequentially distills expert-level reasoning trajectories into a target model under a structured curriculum, effectively circumventing the challenges associated with online reinforcement learning, such as reward design and exploration.

The process leverages a powerful, \textit{frozen} \textbf{Strategy Model} as an expert demonstrator. For a given clinical case $P$, it generates a multitude of candidate reasoning steps. These candidates are rigorously evaluated by the \textbf{T-Eval} framework, which acts as a principled reward function, assigning quality scores based on 
the integrity of specific Toulmin components. The highest-scoring candidates are systematically selected and fused into optimal, structured reasoning trajectories $C^{(k)}$ for each curriculum stage $k$.

A target model $M_\theta$, initialized from a base pre-trained language model, is then trained to imitate these curated trajectories through supervised fine-tuning. The training proceeds sequentially across stages $k = 1, 2, 3$. At each stage, the model learns to generate the corresponding optimal trajectory $C^{(k)}$ by minimizing the negative log-likelihood objective:
\begin{equation}
\mathcal{L}^{(k)}(\theta) = 
\mathbb{E}_{(P, C^{(k)}) \sim \mathcal{D}^{(k)}} 
\Big[ -\log p(C^{(k)} \mid P; \theta) \Big]
\end{equation}
Here, $\mathcal{D}^{(k)}$ denotes the stage-$k$ curriculum dataset consisting of paired clinical cases and the selected/fused optimal trajectories $C^{(k)}$ constructed by the strategy model and T-Eval.

Crucially, the model parameters are updated iteratively, with the model at stage $k$ being initialized from the parameters of the stage $k - 1$ model:
\begin{equation}
\begin{aligned}[t]
\theta^{(k)} = \arg\min_{\theta} \mathcal{L}^{(k)}(\theta),  
\quad \text{with } \\ \theta^{(k)} \text{ initialized from } \theta^{(k-1)}
\end{aligned}
\end{equation}

This iterative refinement, where $\theta^{(0)}$ represents the base model, ensures that competencies acquired in earlier, foundational stages are preserved and built upon, thereby progressively instilling a robust and transparent diagnostic reasoning capability.

\subsubsection{Stage 1: Factual Grounding and Hypothesis Generation}

The first stage of the curriculum targets the establishment of foundational clinical reasoning capabilities: comprehensive data extraction and systematic hypothesis generation. The objective is to produce an initial structured output $C^{(1)}$ that integrates two core Toulmin components—the clinical data ($D$) and a preliminary differential diagnosis ($R$).

To construct the training dataset $\mathcal{D}^{(1)}$, we employ a trajectory collection process. For each clinical case $P$, the strategy model $M_{\text{strategy}}$ generates candidate sets $\{C^{i}_1\}$ for data extraction and $\{C^{j}_2\}$ for differential diagnosis, following respective instructional prompts. Each candidate is evaluated by the T-Eval framework, yielding quality scores $S_D(C^{i}_1)$ and $S_R(C^{j}_2)$. The optimal candidates, selected via $C^{*}_1 = \arg\max_{c^{i}_1} S_D(C^{i}_1)$ and $C^{*}_2 = \arg\max_{c^{j}_2} S_R(C^{j}_2)$, are subsequently fused by the strategy model into a coherent, composite output $C^{(1)}$. This fusion ensures logical continuity between the extracted evidence and the generated hypotheses, forming an optimal Stage 1 reasoning trajectory. The resulting dataset is defined as:
\begin{equation}
\mathcal{D}^{(1)} = \{ (P_i, C^{(1)}_i) \}^{N}_{i=1}.
\end{equation}

The model $M_\theta$ is then initialized with the base parameters $\theta^{(0)}$ and fine-tuned on $\mathcal{D}^{(1)}$ to distill this structured reasoning capability. The stage concludes with the optimization:
\begin{equation}
\begin{aligned}[t]
\theta^{(1)} = \arg\min_{\theta} \mathbb{E}_{(P, C^{(1)}) \sim \mathcal{D}^{(1)}} 
\Big[- \log p(C^{(1)} \mid P;\\ \theta)\Big], 
\quad \text{with } \theta^{(1)} \text{ initialized from } \theta^{(0)},
\end{aligned}
\end{equation}
yielding a model proficient in transforming an unstructured clinical narrative into a solid foundation for subsequent diagnostic argumentation.

This stage instills the fundamental clinical discipline of separating objective observation from interpretation, mirroring the training of medical students who are first taught to meticulously gather facts before forming diagnostic hypotheses.

\subsubsection{Stage 2: Argumentative Justification and Critical Refutation}

Building upon the foundational outputs of Stage 1, the second curriculum stage aims to develop the model's capacity for deep, causal justification and critical evaluation of competing diagnoses. This stage focuses on generating the extended output $C^{(2)}$, which incorporates the Warrant ($W$) for the primary diagnosis and the Backing ($B$) for refuting alternatives, integrated with the prior output $C^{(1)}$.

The data collection protocol for Stage 2 extends the established methodology. For each case $P$, the strategy model is prompted to generate candidate warrants $\{C^{p}_3\}$ and candidate backing rationales $\{C^{q}_4\}$, conditioned on the optimal Stage 1 output $C^{(1)}$. These candidates are evaluated by T-Eval, receiving scores $S_W(C^{p}_3)$ and $S_B(C^{q}_4)$, respectively. The optimal candidates $C^{*}_3$ and $C^{*}_4$ are selected based on these scores. A critical fusion step is then performed, wherein the strategy model synthesizes $C^{(1)}$, $C^{*}_3$, and $C^{*}_4$ into a unified and logically coherent argument $C^{(2)}$. This composite output represents a complete diagnostic justification up to this point. The dataset for this stage is constructed as an augmentation of the previous one:
\begin{equation}
\mathcal{D}^{(2)} = \mathcal{D}^{(1)} \cup \{ (P_i, C^{(2)}_i) \}^{N}_{i=1}.
\end{equation}

The distillation objective for Stage 2 is to fine-tune the Stage 1 model to now generate the more complex output $C^{(2)}$. The optimization leverages the parameters $\theta^{(1)}$ as the starting point, ensuring retention of Stage 1 capabilities while learning new skills:
\begin{equation}
\begin{aligned}[t]
\theta^{(2)} = \arg\min_{\theta} 
\mathbb{E}_{(P, C^{(2)}) \sim \mathcal{D}^{(2)}} 
\Big[- \log p(C^{(2)} \mid P; \\ \theta)\Big], 
\quad \text{with } \theta^{(2)} \text{ initialized from } \theta^{(1)}.
\end{aligned}
\end{equation}

This progression from fact collection to critical reasoning reflects the natural advancement in clinical training, where trainees evolve from passive information gatherers to active, analytical diagnosticians capable of defending their conclusions.

\subsubsection{Stage 3: Synthesis and Qualified Conclusion}

The final stage cultivates the expert-level skill of diagnostic synthesis and calibration, requiring the model to integrate all preceding analysis into a definitive yet appropriately qualified conclusion. A key innovation is the mandatory evidence-based revision mechanism, which instills intellectual honesty and meta-cognitive awareness, namely the ability to recognize and correct one’s own diagnostic errors based on conflicting evidence. The target output is the complete diagnostic argument $C^{(3)}$, which formally integrates all components from previous stages along with the final qualified claim $(Q, Y)$ and, when applicable, an evidence-based revision rationale $\Delta$. The complete argument can thus be represented as $C^{(3)} = C^{(2)} \cup \{Q, Y, \Delta \cdot \mathbb{I}_{\mathrm{rev}}\},$ where $\mathbb{I}_{\mathrm{rev}} \in \{0,1\}$ indicates whether the final diagnosis revises the initial top candidate.

The trajectory collection for Stage 3 completes the reasoning process. Conditioned on the full contextual argument $C^{(2)}$, the strategy model generates candidate final diagnoses and qualifiers $\{C_5^r\}$. T-Eval assesses these candidates with emphasis on claim correctness ($S_Y$) and qualifier quality ($S_Q$), where the latter is activated when the final diagnosis revises the initial top candidate. The optimal candidate $C_5^*$ is fused with $C^{(2)}$ to produce the final trajectory $C^{(3)}$, forming the ultimate training dataset:
\begin{equation}
\mathcal{D}^{(3)} = \mathcal{D}^{(2)} \cup \{(P_i, C^{(3)}_i)\}_{i=1}^{N}.
\end{equation}

The model distillation objective for Stage 3 fine-tunes the Stage 2 model to generate the complete argument $C^{(3)}$ from the clinical case $P$:
\begin{equation}
\begin{aligned}[t]
\theta^{(3)} = \arg\min_{\theta} \mathbb{E}_{(P, C^{(3)}) \sim \mathcal{D}^{(3)}} 
\Big[ -\log p\big(C^{(3)} \mid P;\\ \theta\big) \Big], 
\quad \text{with } \theta^{(3)} \text{ initialized from } \theta^{(2)}.
\end{aligned}
\end{equation}
The resulting model $M_{\theta^{(3)}}$ embodies the culmination of the CGCL pipeline, capable of executing end-to-end diagnostic reasoning that transparently exhibits all Toulmin components through structured reasoning trajectories.

\begin{table*}[ht]
\centering
\fontsize{8pt}{8pt}\selectfont
\setlength{\tabcolsep}{15pt}
\renewcommand{\arraystretch}{1.0}
\begin{tabular}{@{}c|c|c|c|c|c@{}}
\toprule
\rowcolor{gray!50} 
\multicolumn{6}{c}{\textbf{MedCaseReasoning}} \\ 
\midrule
\rowcolor{gray!25} 
\multicolumn{2}{c|}{\textbf{Method}}
& \multicolumn{2}{c|}{\textbf{Qwen2.5-7B-Instruct}} 
& \multicolumn{2}{c}{\textbf{Qwen2.5-3B-Instruct}} \\
\midrule
Category & Variant
& \textbf{Accuracy }($\uparrow$) & \textbf{TrustScore} ($\uparrow$)
& \textbf{Accuracy} ($\uparrow$) & \textbf{TrustScore} ($\uparrow$) \\
\midrule
\multirow{3}{*}{Prompt}
& Vanilla & 22.37 & - & 20.61 & - \\
& Think \& Answer & 25.68 & 62.43 & 23.47 & 61.32 \\
& \textbf{Trust-Think \& Answer} & 26.31 & 65.52 & 23.42 & 63.28 \\
\midrule
\multirow{3}{*}{SFT}
& SFT-GT & 31.72 & 63.74 & 28.16 & 62.38 \\
& SFT-CoT & 32.08 & 60.79 & 28.27 & 59.83 \\
& \textbf{SFT-CGCL(Stage3)} & 30.12 & 70.78 & 26.58 & 68.67 \\
\midrule
\multirow{2}{*}{RL}
& DPO & 32.24 & 71.18 & 28.59 & 69.27 \\
& GRPO & \textbf{32.81} & \textbf{73.12} & \underline{28.90} & \underline{71.25} \\
\midrule
\multirow{1}{*}{Ours}
&\textbf{ CGCL} & \underline{32.63} & \underline{72.84} & \textbf{28.95} & \textbf{71.30} \\
\midrule
\rowcolor{gray!25} 
\multicolumn{2}{c|}{\textbf{Method}}
& \multicolumn{2}{c|}{\textbf{LLaMA3.1-8B-Instruct}} 
& \multicolumn{2}{c}{\textbf{LLaMA3.2-3B-Instruct}} \\
\midrule
Category & Variant
& \textbf{Accuracy }($\uparrow$) & \textbf{TrustScore} ($\uparrow$)
& \textbf{Accuracy} ($\uparrow$) & \textbf{TrustScore} ($\uparrow$) \\
\midrule
\multirow{3}{*}{Prompt}
& Vanilla & 22.46 & - & 20.57 & - \\
& Think \& Answer & 25.73 & 62.58 & 23.51 & 61.29 \\
& \textbf{Trust-Think \& Answer} & 26.38 & 65.47 & 23.43 & 63.36 \\
\midrule
\multirow{3}{*}{SFT}
& SFT-GT & 31.79 & 63.81 & 28.12 & 62.49 \\
& SFT-CoT & 32.14 & 60.86 & 28.33 & 59.91 \\
& \textbf{SFT-CGCL(Stage3)} & 30.07 & 70.92 & 26.64 & 68.59 \\
\midrule
\multirow{2}{*}{RL}
& DPO & 32.29 & 71.22 & 28.63 & 69.31 \\
& GRPO & \textbf{32.83} & \textbf{73.24} & \textbf{28.95} & \underline{71.30} \\
\midrule
\multirow{1}{*}{Ours}
&\textbf{ CGCL} & \underline{32.67} & \underline{72.89} & \underline{28.90} & \textbf{71.35} \\
\midrule
\end{tabular}
\caption{Overall Performance Comparison on MedCaseReasoning Benchmark. Paired significance tests and 95\% confidence intervals for the key CGCL comparisons are reported in Appendix \ref{app:exp:stats} (Table~\ref{tab:significance}).}
\label{tab1}
\end{table*}

\subsection{T-Eval: A Toulmin-Based Evaluation Framework}

\subsubsection{Evaluation Dimensions and Scoring Metrics}
The T-Eval framework provides a principled methodology for quantifying the quality of diagnostic reasoning by directly operationalizing the Toulmin components established in Section \ref{Toulmin Model}. We define six evaluation dimensions that correspond to the core elements of our clinical argumentation mapping:
\textbf{Data Score ($S_D$)}: Measures completeness and accuracy of \textbf{Clinical Evidence} extraction.
\textbf{Warrant Score ($S_W$)}: Assesses medical plausibility of \textbf{Pathophysiological Rationale}.
\textbf{Backing Score ($S_B$)}: Quantifies appropriate use of \textbf{Principled Justification}.
\textbf{Rebuttal Score ($S_R$)}: Evaluates thoroughness of \textbf{Differential Diagnosis} analysis.
\textbf{Qualifier Score ($S_Q$)}: Assesses the appropriateness and justification of the diagnostic \textbf{Certainty Assessment}. 
\textbf{Claim Score ($S_Y$):} Measures the correctness of the final diagnosis.

Each dimension score $S_c^{(i)}$ is rated on a 1--5 Likert scale. We first normalize it to $\tilde{S}_c^{(i)} = (S_c^{(i)} - 1)/4 \in [0, 1]$. The overall \textbf{T-Eval TrustScore} is then computed as the average normalized reasoning quality across the five Toulmin components, scaled to $[0, 100]$:
\begin{equation}
\label{eq:TrustScore}
\begin{aligned}[t]
\text{T-Eval TrustScore} = \frac{100}{5N} \sum_{i=1}^N \sum_{c\in\{D,W,B,R,Q\}} \tilde{S}_c^{(i)}.
\end{aligned}
\end{equation}
Relation to prior TrustScore. The term TrustScore has been used in prior work \citep{zheng2024trustscore} as a reference-free trustworthiness metric for general LLM responses. In contrast, our metric aggregates rubric-based judgments over Toulmin components tailored to clinical diagnostic reasoning. To avoid ambiguity, we refer to our metric as T-Eval TrustScore throughout.
We define diagnostic \textbf{Accuracy} using the T-Eval claim score $S_Y\in\{1,\dots,5\}$.
Specifically, we normalize the claim score as $\tilde{S}_Y=(S_Y-1)/4\in[0,1]$ and aggregate over cases:
\begin{equation}
\label{eq:acc_teval}
{Accuracy}=\frac{100}{N}\sum_{i=1}^{N}\tilde{S}_Y^{(i)}.
\end{equation}
This yields a calibrated 0--100 measure of claim correctness derived from rubric-guided judging, consistent with our component scoring. We compute TrustScore over $\{D,R,W,B,Q\}$ to measure reasoning quality \emph{orthogonally} to end-task correctness;  the final claim is evaluated separately via the T-Eval claim score $S_Y$ (reported as Accuracy in Eq.~\ref{eq:acc_teval}).

\subsubsection{Implementation with Calibrated LLM Judges}
This implementation enables scalable and reproducible assessment via rubric-guided prompting and a multi-judge consensus mechanism (three independent LLM instances with aggregation and outlier handling). We emphasize that the main T-Eval scores are produced automatically by LLM judges.\\
\textbf{Clinician validation.} We validated the automated T-Eval TrustScore against expert judgment by sampling cases from the MedCaseReasoning test set. Board-certified clinicians, blinded to the source, rated reasoning traces using the same Toulmin-aligned 1–5 Likert rubric. We averaged these ratings to form a Clinician TrustScore and calculated its correlation (Spearman’s $\rho$) with the T-Eval TrustScore, alongside inter-rater reliability (ICC). See Appendix~\ref{app:exp:clinician} for full study protocols.

\begin{figure*}[ht] 
  \centering
  \begin{subfigure}[b]{0.48\textwidth} 
    \centering
    \includegraphics[width=\linewidth]{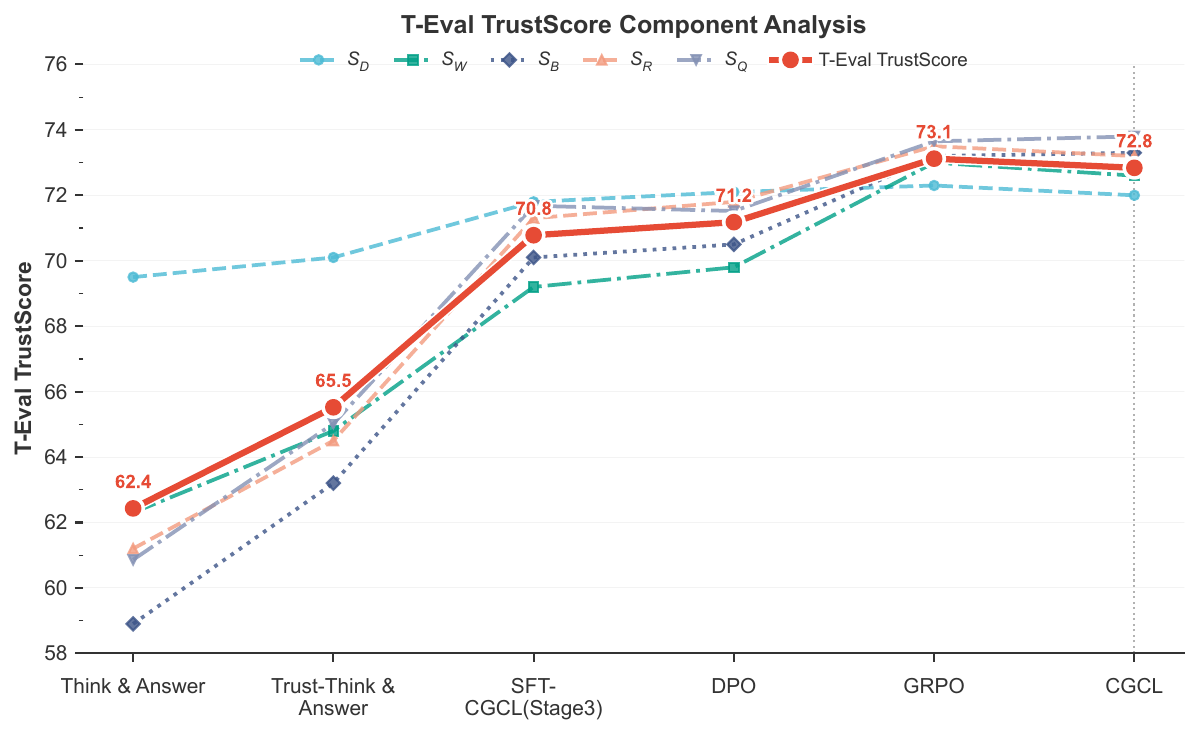}
    \caption{T-Eval TrustScore Component Analysis.}
    \label{fig4}
  \end{subfigure}
  \hfill 
  \begin{subfigure}[b]{0.48\textwidth}
    \centering
    \includegraphics[width=\linewidth]{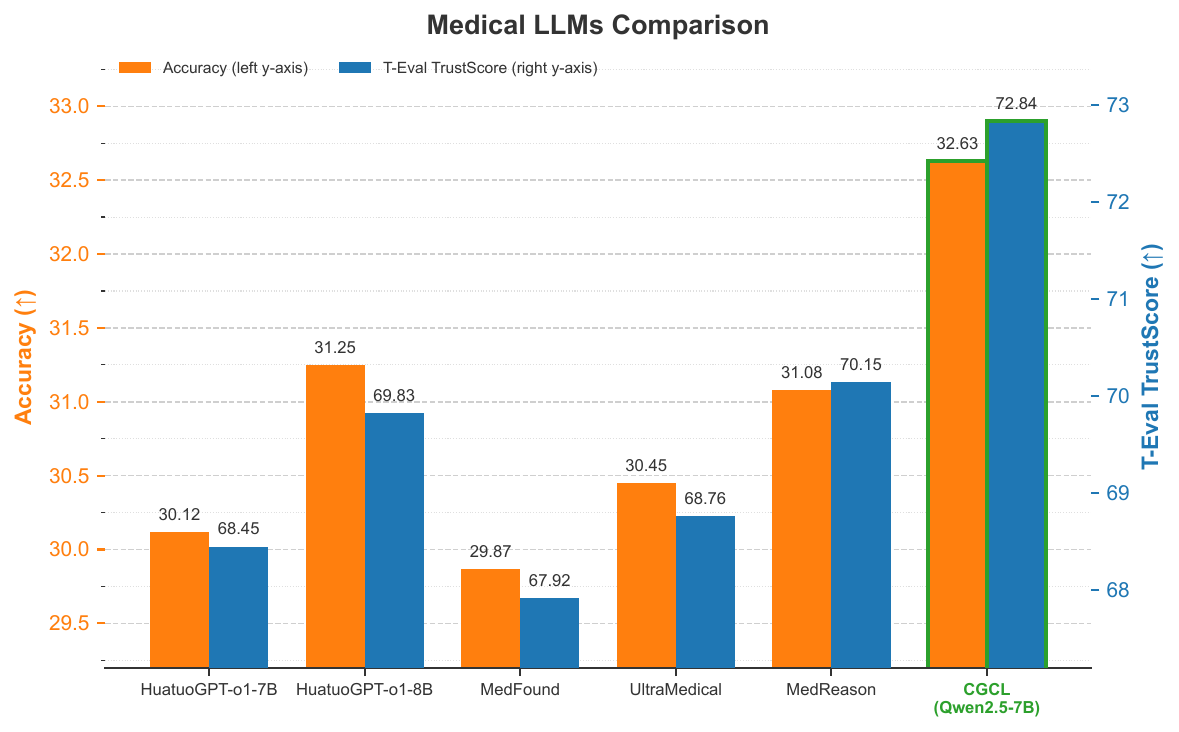}
    \caption{Comparison with Medical LLMs.}
    \label{fig3}
  \end{subfigure}
  
  \caption{Performance and trustworthiness analysis.}
  \label{fig:combined_wide}
\end{figure*}

\section{Experiments}
\subsection{Experimental Setup}

\textbf{Datasets.} We conduct our experiments on MedCaseReasoning\footnote{\url{https://huggingface.co/datasets/zou-lab/MedCaseReasoning}} \citep{wu2025medcasereasoning}, a publicly available complex medical diagnostic reasoning benchmark. This open-access dataset comprises over 14,000 clinical diagnostic cases sourced from more than 800 medical journals across 30$+$ medical specialties. For evaluation, we use the official 897-case test split of MedCaseReasoning, which contains challenging diagnostic scenarios.\\
\textbf{Baselines.} We compare CGCL against a comprehensive set of baseline approaches, categorized into four groups:

\textbf{Prompt-based.} Vanilla Direct answer generation without intermediate reasoning steps. Think \& Answer Standard chain-of-thought prompting where models reason within \texttt{<think>} tags before producing final answers in \texttt{<answer>} tags. Trust-Think \& Answer Our enhanced reasoning approach that employs comprehensive argumentation to support step-by-step diagnostic reasoning.

\textbf{SFT-based.} SFT-GT Supervised fine-tuning using ground-truth answers only. SFT-CoT Supervised fine-tuning with common chain-of-thought reasoning traces. SFT-CGCL (Stage3) Learns the complete Toulmin structure in a single stage without progressive curriculum learning.

\textbf{RL-based.} DPO \citep{rafailov2023direct} Direct Preference Optimization using contrastive pairs of correct vs. incorrect reasoning trajectories. GRPO \citep{shao2024deepseekmath} Group Relative Policy Optimization with trajectory-level correctness rewards.

\textbf{Medical LLMs.} We include comparisons with state-of-the-art medical and reasoning-specialized models including HuatuoGPT-o1 \citep{chen2024huatuogpt}, MedFound \citep{liu2025generalist}, UltraMedical \citep{zhang2024ultramedical}, and MedReason \citep{wu2025medreason} to establish comprehensive performance benchmarks.
For reproducibility, we provide algorithmic details (Appendix~\ref{app:B}), training recipes, and hyperparameters (Appendix~\ref{app:C}), and prompt templates and judge rubrics (Appendix~\ref{app:D}).\\
\textbf{Evaluation Metrics.} We employ two primary evaluation metrics: \textbf{Accuracy} and \textbf{T-Eval TrustScore}. Diagnostic Accuracy measures claim correctness derived from the T-Eval claim score $S_Y$ (Eq.~\ref{eq:acc_teval}), while T-Eval TrustScore summarizes reasoning quality over $\{D,R,W,B,Q\}$ (Eq.~\ref{eq:TrustScore}).\\
\textbf{Implementation Details.} Our experiments are conducted across diverse model architectures and scales: LLaMA3.1-8B, LLaMA3.2-3B \citep{dubey2024llama}, Qwen2.5-7B/3B \citep{qwen2025qwen25technicalreport}. The strategy model for trajectory generation employs DeepSeek-R1 \citep{guo2025deepseek}. For training-based approaches, models are fine-tuned on the training splits of each benchmark and evaluated on the respective test sets. Prompt-based methods are evaluated directly in a zero-shot setting. \\
\textbf{Compute and training cost.} To substantiate the resource-efficiency claim, we report GPU-hours, and peak GPU memory for CGCL, DPO, and GRPO under matched compute budgets and batch settings (Table~\ref{tab:exp:compute}; Appendix~\ref{app:exp:compute}).

\subsection{Main Results}
\textbf{Performance Comparison.} Table~\ref{tab1} summarizes the performance of CGCL and all baselines across model sizes. Overall, CGCL yields consistently higher reasoning quality than prompting and standard SFT baselines while maintaining competitive diagnostic accuracy. Compared with GRPO, CGCL is comparable across scales: it is marginally higher on 3B models ($+$0.05 TrustScore) and slightly lower on 7B/8B models. We further report paired significance tests and confidence intervals for these small gaps in Appendix~\ref{app:exp:stats}.\\
\textbf{T-Eval TrustScore Analysis.} The component-level analysis in Figure~\ref{fig4} reveals CGCL's balanced improvements across Toulmin components, with leading performance in backing quality ($S_B$ = 73.3) and qualifier quality ($S_Q$ = 73.8). Compared to prompt-based methods, CGCL achieves significant T-Eval TrustScore improvements, with notable gains in warrant and backing qualities. The performance advantage over SFT-CGCL (Stage3) (72.84 vs. 70.78) underscores the importance of progressive curriculum. Since the SFT-CGCL(Stage3) baseline is trained on the exact same set of high-quality trajectories as the full CGCL model but without staging, the observed gain is directly attributable to the curriculum structure rather than data quality alone. While GRPO maintains slight edges in data, warrant, and rebuttal qualities on larger models, CGCL's superior performance on smaller models and in critical reasoning components validates its efficiency and balanced reasoning capabilities.
\subsection{Analysis}
\label{analysis}
\textbf{Comparison with Medical LLMs.} Figure \ref{fig3} compares CGCL to recent medical and reasoning-specialized LLMs on MedCaseReasoning under the same prompting and scoring pipeline. Across both 3B and 8B backbones, CGCL achieves the best or competitive T-Eval TrustScore while maintaining similar diagnosis Accuracy. These results indicate that explicitly training Toulmin-structured argumentation can improve the organization and justification of diagnostic reasoning even when starting from a general-purpose backbone. We emphasize that this comparison is limited to MedCaseReasoning and our evaluation protocol, and does not measure broader clinical knowledge coverage or real-world safety. To further assess generalization beyond MedCaseReasoning, Appendix~\ref{app:medfound_ood} reports zero-shot OOD evaluation on the external \textsc{MedFound} benchmark. The overall ranking trend remains consistent: CGCL outperforms prompting and standard SFT baselines and remains competitive with RL-based methods.\\
\begin{table}[h]
\centering
\renewcommand{\arraystretch}{1.0}
\resizebox{\linewidth}{!}{%
    \setlength{\tabcolsep}{6pt} 
    \begin{tabular}{@{}lcc@{}}
    \toprule
    \rowcolor{gray!15} 
    \multicolumn{3}{c}{\textbf{Clinician Validation Performance}} \\
    \midrule
    Method & Clinician TrustScore $\uparrow$ & Dx Acc. (\%) $\uparrow$ \\
    \midrule
    Trust-Think \& Answer & $51.5$ & 46.7 \\
    SFT-CoT               & $54.5$ & 48.3 \\
    GRPO                  & $61.0$ & 51.7 \\
    \textbf{CGCL (Ours)}  & \boldmath$71.0$ & \textbf{53.3} \\
    \bottomrule
    \end{tabular}%
}
\caption{Clinician validation on $N_{\text{clin}}=50$ randomly sampled test cases.}
\label{tab:clinician_main}
\end{table}
\textbf{Clinician validation.} Clinicians rank CGCL highest in the human evaluation, with a clinician TrustScore of $71.0 \pm 0.7$ and diagnosis accuracy of $53.3$ (Table~\ref{tab:clinician_main}). Compared to GRPO ($61.0 \pm 0.8$), CGCL yields a +10.0 gain in clinician TrustScore, mirroring the main T-Eval ranking and supporting that the improvements are not artifacts of LLM judging. Inter-rater reliability is moderate (ICC$(2,k)=0.63$), and T-Eval correlates with clinician assessments with Spearman’s $\rho=0.74$, suggesting that T-Eval is a reasonable proxy for expert judgment while leaving room for future improvements in human-aligned evaluation.\\
\begin{table}[h]
\centering
\fontsize{9pt}{9pt}\selectfont
\setlength{\tabcolsep}{12pt}
\renewcommand{\arraystretch}{1.1}
\begin{tabular}{@{}c|c|c@{}}
\toprule
\rowcolor{gray!15} 
\multicolumn{3}{c}{\textbf{Curriculum Stage Ablation}} \\ 
\midrule
Curriculum Stage & Accuracy ($\uparrow$) &TrustScore ($\uparrow$) \\
\midrule
Stage 1 only & 28.45 & 65.32 \\
Stage 1+2 only & 30.67 & 69.84 \\
\textbf{CGCL (full)} & \textbf{32.63} & \textbf{72.84} \\
\bottomrule
\end{tabular}
\caption{Ablation study of curriculum learning stages. }
\label{tab:ablation}
\end{table}
\textbf{Ablation Studies.} The curriculum stage ablation in Table \ref{tab:ablation} validates the progressive nature of our training approach. Stage 1 alone, focusing on factual grounding and hypothesis generation, achieves moderate performance (65.32 T-Eval TrustScore). Adding Stage 2, which introduces argumentative justification and critical refutation, provides a substantial 4.52 point T-Eval TrustScore improvement. The complete three-stage curriculum yields the best results, with an additional 3.00 point gain, demonstrating that synthesis and qualification capabilities are essential for optimal diagnostic reasoning. This monotonic improvement across stages confirms that each curriculum phase builds upon previously acquired skills, and compressing the training into fewer stages compromises reasoning quality.

\section{Conclusion}
We propose CGCL and the T-Eval evaluation framework, which enhance the transparency and trustworthiness of large language models in clinical diagnostic reasoning using the Toulmin model of argumentation. Experiments on MedCaseReasoning demonstrate that CGCL consistently improves T-Eval TrustScore compared to prompting and standard SFT baselines, while achieving accuracy comparable to strong RL-based approaches, particularly with smaller models. Ablation studies validate the effectiveness of our progressive curriculum design, showing continuous improvements in reasoning quality across the three-stage training process. Our results suggest that, compared to medical LLMs, structured reasoning training can partly compensate for the lack of domain-specific pre-training on MedCaseReasoning. Looking ahead, we aim to explore richer supervision through multi-expert trajectory distillation and extend the framework to incorporate multi-modal clinical data. This will further enhance the model’s practicality and reliability in real-world clinical environments, while maintaining its strong theoretical foundation for trustworthy clinical decision support systems.

\section*{Limitations}
While the proposed CGCL framework demonstrates significant improvements in clinical diagnostic reasoning, several limitations should be addressed. First, the performance of the framework is inherently constrained by the quality and diversity of the trajectory generation model. This may not fully capture the complete range of clinical reasoning patterns. Second, although our evaluation is comprehensive, it primarily focuses on diagnostic accuracy and structured reasoning quality, without testing the model in real-time clinical workflows or integrating it with existing clinical decision support systems. Third, the current implementation relies on synthetic trajectory generation and distillation, which may not fully replicate the nuanced decision-making processes of human clinical experts across all medical specialties. While the Toulmin schema and T-Eval penalize inconsistent arguments, models can still arrive at correct final diagnoses despite flawed intermediate reasoning. A targeted qualitative audit to identify failure modes such as ‘correct answer, wrong reason’ at intermediate steps  will be an important next step. The system is intended as decision support and must be used under clinician supervision; it is not a substitute for professional medical judgment. Finally, the computational requirements of the multi-stage training process, particularly trajectory generation and fusion, present challenges for rapid deployment in resource-constrained clinical settings. These limitations point to key directions for future research and further refinement of our approach.




\bibliography{custom}

\appendix

\begin{table*}[t]
    \centering
    \small
    \setlength{\tabcolsep}{6pt}
    \renewcommand{\arraystretch}{1.18}
    \caption{Notation, data objects, and evaluation scores used in this work.}
    \label{tab:app-notation}
    \begin{tabular}{p{0.16\textwidth} p{0.79\textwidth}}
        \toprule
        \textbf{Symbol} & \textbf{Description} \\
        \midrule

        \multicolumn{2}{l}{\textit{\textbf{I. Clinical data \& model objects}}} \\
        $P$ & \textbf{Clinical case}: unstructured narrative containing symptoms, signs, labs, and imaging findings. \\
        $M_{\text{strategy}}$ & \textbf{Strategy model}: frozen LLM used to generate candidate reasoning components. \\
        $\{C^{Z}_{i,j}\}_{j=1}^{K_Z}$ & \textbf{Candidate set}: candidates for component $Z$, where $Z \in \{D,R,W,B,Q,Y\}$. \\

        \midrule
        \multicolumn{2}{l}{\textit{\textbf{II. Toulmin-aligned reasoning components}}} \\
        $D$ & \textbf{Data}: objective clinical evidence extracted from $P$. \\
        $R$ & \textbf{Rebuttal}: differential diagnoses capturing alternative hypotheses and counter-arguments. \\
        $W$ & \textbf{Warrant}: pathophysiological rationale connecting evidence ($D$) to hypotheses. \\
        $B$ & \textbf{Backing}: principled justification (guidelines, medical knowledge) supporting $W$. \\
        $Q$ & \textbf{Qualifier}: uncertainty calibration and conditions under which the claim holds. \\
        $Y$ & \textbf{Claim}: final diagnosis (ultimate clinical judgment). \\

        \midrule
        \multicolumn{2}{l}{\textit{\textbf{III. Curriculum learning \& evaluation}}} \\
        $C^{(k)}$ & \textbf{Stage-$k$ trajectory} with progressive structure:
        \begin{itemize}[leftmargin=1.2em, topsep=0pt, itemsep=0pt, parsep=0pt]
            \item Stage 1: $C^{(1)} = (D, R)$
            \item Stage 2: $C^{(2)} = (D, R, W, B)$
            \item Stage 3: $C^{(3)} = (D, R, W, B, Q, Y)$
        \end{itemize} \\

        $\mathcal{D}^{(k)}$ & \textbf{Curriculum dataset}: $\mathcal{D}^{(k)}=\{(P_i, C^{(k)}_i)\}_{i=1}^{N}$, containing fused optimal stage-$k$ trajectories. \\

        $S_D$ & \textbf{Data score}: Likert (1--5) rating of evidence completeness, accuracy, and grounding in $P$. \\
        $S_R$ & \textbf{Rebuttal score}: Likert (1--5) rating of differential coverage and quality of counter-arguments. \\
        $S_W$ & \textbf{Warrant score}: Likert (1--5) rating of mechanistic plausibility linking $D$ to leading hypotheses. \\
        $S_B$ & \textbf{Backing score}: Likert (1--5) rating of principled justification quality (clinical rules/knowledge) supporting $W$, penalizing fabricated claims. \\
        $S_Q$ & \textbf{Qualifier score}: Likert (1--5) rating of uncertainty calibration, missing-information awareness, and conditions/limitations stated in $Q$. \\
        $S_Y$ & \textbf{Claim score}: Likert (1--5) rating of the final diagnosis $Y$, primarily reflecting correctness and consistency with the case evidence. \\

        $\tilde{S}_Z$ & \textbf{Normalized score}: $\tilde{S}_Z=(S_Z-1)/4 \in [0,1]$ for $Z \in \{D,R,W,B,Q,Y\}$. \\

        $\text{TrustScore}$ & \textbf{T-Eval TrustScore}: average of normalized component scores (over the evaluated components), scaled to $[0,100]$ (Eq.~\ref{eq:TrustScore}). \\

        $\text{Accuracy}$ & \textbf{Diagnostic accuracy:} aggregated claim correctness derived from $S_Y$ (Eq.~\ref{eq:acc_teval}).\\
        
        \bottomrule
    \end{tabular}
\end{table*}

\section{Problem Formulation and Toulmin Model Instantiation}
\label{app:toulmin}
\subsection{The Toulmin Model of Argumentation}
The Toulmin model provides a robust framework for analyzing informal arguments by decomposing them into six fundamental components: (1) \textbf{Claim (Y)}: the final assertion or conclusion; (2) \textbf{Data (D)}: the facts and evidence used to support the claim; (3) \textbf{Warrant (W)}: the logical bridge connecting the data to the claim; (4) \textbf{Backing (B)}: the general principles or authority that reinforces the warrant; (5) \textbf{Qualifier (Q)}: the modality or degree of certainty associated with the claim; and (6) \textbf{Rebuttal (R)}: the consideration and refutation of counter-arguments or alternative claims.

\subsection{Clinical Diagnostic Reasoning as Structured Argumentation}
We propose that the clinical diagnostic process is fundamentally an exercise in structured argumentation. To operationalize the Toulmin model for our CGCL framework, we establish the following precise mapping from its components to core clinical reasoning concepts:\\
\textbf{Data (D)}: \textbf{Clinical Evidence} : The objective facts extracted from the patient's presentation, including symptoms, signs, laboratory results, and imaging findings. This constitutes the foundational evidence for the diagnostic argument.\\
\textbf{Rebuttal (R)}: \textbf{Differential Diagnosis} : The systematic consideration and ranked listing of alternative diagnostic hypotheses. The process of comparing and prioritizing these competing claims embodies the rebuttal component by arguing against the exclusive validity of any single hypothesis. \\
\textbf{Warrant (W)}: \textbf{Pathophysiological Rationale} : The logical bridge and mechanistic explanation that connects the clinical evidence to a specific diagnostic hypothesis. It provides the causal reasoning, typically grounded in disease mechanisms, for why the evidence supports the claim. \\
\textbf{Backing (B)}: \textbf{Principled Justification} : The established medical knowledge, clinical guidelines, or diagnostic criteria that authorize and validate the warrants. This component provides the authoritative foundation for both supporting the primary diagnosis and refuting alternatives. \\
\textbf{Qualifier (Q)}: \textbf{Certainty Assessment} : The explicit evaluation of diagnostic confidence and, when applicable, the rationale for any revision from initial hypotheses. This component reflects the calibrated nature of clinical decision-making and acknowledges diagnostic uncertainty.\\
\textbf{Claim (Y)}: \textbf{Final Diagnosis} : The final diagnostic determination that synthesizes all available evidence and reasoning. This represents the definitive clinical judgment based on the complete analytical process.
This structured mapping ensures that each abstract Toulmin component corresponds to a concrete, clinically-meaningful reasoning step, providing a transparent foundation for our staged curriculum and evaluation framework. Because MedCaseReasoning requires \textbf{free-form} diagnostic conclusions rather than selecting from a fixed option list, exact-string-match accuracy is ill-defined (e.g., synonyms, abbreviations, clinically equivalent variants, or multi-condition diagnoses). We therefore operationalize \textbf{Accuracy} using T-Eval's rubric-guided \textbf{claim correctness} score $S_Y \in \{1, \ldots, 5\}$, where 5 denotes an exact match and 4 denotes a clinically equivalent variant. We normalize $S_Y$ to $[0,1]$ and report the dataset-level mean scaled to $[0,100]$ as Accuracy (Eq.~\ref{eq:acc_teval}), making it consistent with our component-wise scoring protocol.

\section{Related Work}

\textbf{Medical Large Models and Diagnostic Reasoning.} Research on medical large language models has primarily advanced along two directions: 1) developing domain-specific architectures and pre-training strategies, such as ClinicalBERT and BioBERT, which are continually pre-trained on clinical texts \citep{huang2019clinicalbert,lee2020biobert}; and 2) enhancing the performance of general-purpose large models on medical question-answering tasks through instruction fine-tuning, as seen with Med-PaLM and ChatDoctor \citep{singhal2023large,li2023chatdoctor}. These studies typically rely on standardized benchmarks like MedQA and PubMedQA for evaluation, employing a paradigm centered on answer-matching \citep{jin2019pubmedqa,pal2022medmcqa}. This approach, however, struggles to detect instances of "correct answers from flawed reasoning." Although recent work has attempted to elicit reasoning steps via Chain-of-Thought (CoT) prompting, the resulting free-form reasoning chains lack structural constraints, and their logical rigor and medical reliability cannot be effectively evaluated or guaranteed \citep{wei2022chain,wang2022self,zhou2022least,yao2023tree}.\\
\textbf{The Toulmin Model of Argumentation.} The Toulmin model serves as a classic and robust theoretical framework for analyzing informal arguments by deconstructing them into six core structured components: Claim, Data (Grounds), Warrant, Backing, Qualifier, and Rebuttal \citep{erduran2004tapping}. This model has seen widespread application in fields such as computational argumentation and educational assessment, where it has been used, for instance, to automatically analyze the argumentation quality of student essays \citep{lawrence2020argument}. In the AI domain, it has also been explored as an explainability tool for parsing the decision logic of intelligent systems \citep{vassiliades2021argumentation}. However, while Toulmin-style argumentation has been discussed in prior medical AI literature \citep{fejer2022survey}, existing work has predominantly focused on using the model for post-hoc analysis of arguments rather than as an a priori structural constraint to be systematically integrated into a model's training process to directly shape its reasoning capabilities. In the more specific setting of long-form clinical diagnostic reasoning, using Toulmin as an explicit training objective together with a scalable evaluation framework remains relatively underexplored.\\
\textbf{Curriculum Learning and Goal-Conditioned Learning.} Curriculum Learning (CL) enhances model training by presenting data or tasks in a sequence of increasing difficulty, a strategy validated for its ability to improve generalization and convergence in tasks like machine translation and visual reasoning \citep{soviany2022curriculum,graves2017automated}. Goal-Conditioned Learning (GCL), on the other hand, guides a model to produce target-oriented outputs by incorporating a goal signal into its input. This paradigm has proven highly effective in reinforcement learning and robotics and is gradually being introduced to sequence generation tasks \citep{yuan2022easy,ghosh2019learning,riedmiller2018learning,dathathri2019plug}. To date, research has largely explored these two paradigms in isolation: CL has centered on data scheduling, while GCL has focused on goal encoding. How to deeply synthesize the staged progression of CL with the explicit guidance of GCL to construct a systematic training pathway for complex reasoning capabilities remains a compelling open research question.

\section{Algorithm Details}
\label{app:B}

\subsection{Notation and Data Objects}
\label{app:algo:notation}

Table~\ref{tab:app-notation} summarizes the notation used in CGCL and T-Eval to support reproducibility and ease cross-referencing with Algorithms~\ref{alg:cgcl}--\ref{alg:teval}.

\subsection{CGCL Trajectory Construction}
\label{app:algo:cgcl}

Algorithm~\ref{alg:cgcl} describes how we construct the stage-wise optimal trajectories using the frozen strategy model and T-Eval.\\
\begin{algorithm}[h]
\small
\caption{CGCL trajectory construction for a single case $P$ (sequential, context-conditioned selection).}
\label{alg:cgcl}
\begin{algorithmic}[1]
\Require Clinical case $P$; strategy model $M_{\text{strategy}}$; evaluator $\textsc{TEval}(\cdot)$;
candidate budgets $\{K_D,K_R,K_W,K_B,K_Q,K_Y\}$; fusion operator $\textsc{Fuse}(\cdot)$.
\Ensure Stage trajectories $C^{(1)}, C^{(2)}, C^{(3)}$.

\Statex \textbf{Notes:} $C_{\text{ctx}}$ denotes the current partial trajectory (context).
\Statex \hspace{\algorithmicindent} $\oplus$ merges a candidate component into $C_{\text{ctx}}$.
\Statex \hspace{\algorithmicindent} $\textsc{TEval}(P, C_{\text{ctx}} \oplus \cdot, \text{dim}=Z)$ returns a Likert score $S_Z \in \{1,\dots,5\}$.

\Function{SelectBest}{$Z, C_{\text{ctx}}, K_Z$} \Comment{$Z \in \{D,R,W,B,Q,Y\}$}
    \State $\{C^Z_j\}_{j=1}^{K_Z} \gets M_{\text{strategy}}(P, C_{\text{ctx}}, \text{prompt}_Z)$
    \For{$j=1$ to $K_Z$}
        \State $S^Z_j \gets \textsc{TEval}(P, C_{\text{ctx}} \oplus C^Z_j, \text{dim}=Z)$
    \EndFor
    \State \Return $C^{Z*} \gets C^Z_{\arg\max_j S^Z_j}$
\EndFunction

\Statex \textbf{Stage 1: construct $(D,R)$}
\State $C^{D*} \gets \Call{SelectBest}{D,\emptyset,K_D}$
\State $C^{R*} \gets \Call{SelectBest}{R,C^{D*},K_R}$
\State $C^{(1)} \gets \textsc{Fuse}(P; C^{D*}, C^{R*})$

\Statex \textbf{Stage 2: augment with $(W,B)$}
\State $C^{W*} \gets \Call{SelectBest}{W,C^{(1)},K_W}$
\State $C^{B*} \gets \Call{SelectBest}{B,C^{(1)} \oplus C^{W*},K_B}$
\State $C^{(2)} \gets \textsc{Fuse}(P; C^{(1)}, C^{W*}, C^{B*})$

\Statex \textbf{Stage 3: augment with $(Q,Y)$}
\State $C^{Q*} \gets \Call{SelectBest}{Q,C^{(2)},K_Q}$
\State $C^{Y*} \gets \Call{SelectBest}{Y,C^{(2)} \oplus C^{Q*},K_Y}$
\State $C^{(3)} \gets \textsc{Fuse}(P; C^{(2)}, C^{Q*}, C^{Y*})$

\State \Return $C^{(1)}, C^{(2)}, C^{(3)}$
\end{algorithmic}
\end{algorithm}\\
\textbf{Fusion operator.} $\textsc{Fuse}$ produces a coherent structured output with a fixed schema (Appendix~\ref{app:prompt-fusion}).
The $\textsc{Fuse}$ operator synthesizes selected components into a coherent, structured output adhering to a strict schema. To ensure integrity, we employ a deterministic post-processing pipeline consisting of three steps: (i) De-duplication to remove redundant clinical evidence; (ii) Consistency Verification between the evidence ($D$) and final diagnosis ($Y$), where a detected conflict triggers a one-time regeneration of the qualifier ($Q$) and diagnosis ($Y$); and (iii) Format Validation to guarantee machine parsability.

\subsection{T-Eval Scoring and Multi-Judge Aggregation}
\label{app:algo:teval}

Algorithm~\ref{alg:teval} details how T-Eval uses multiple independent LLM judges and robust aggregation to score each dimension.
\begin{algorithm}[h]
\small
\caption{T-Eval scoring with multi-judge robust aggregation.}
\label{alg:teval}
\begin{algorithmic}[1]
\Require Case $P$; candidate (partial) trajectory $C$; target dimension $\delta \in \{D,R,W,B,Q,Y\}$;
judges $\{J_m\}_{m=1}^{3}$; rubric $\mathcal{R}_\delta$.
\Ensure Aggregated Likert score $S_\delta(C) \in \{1,2,3,4,5\}$.

\For{$m=1$ to $3$}
    \State Query judge $J_m$ with $(P, C, \mathcal{R}_\delta)$ and obtain $s_m \in \{1,\dots,5\}$.
\EndFor
\State Sort $\{s_1,s_2,s_3\}$ to obtain $s_{(1)} \le s_{(2)} \le s_{(3)}$.
\If{$s_{(3)} - s_{(1)} \ge 3$} \Comment{high disagreement}
    \State $\hat{s} \gets s_{(2)}$ \Comment{use median}
\Else
    \State $\hat{s} \gets \mathrm{round}\!\left(\frac{s_1+s_2+s_3}{3}\right)$
\EndIf
\State \Return $S_\delta(C) \gets \min(5,\max(1,\hat{s}))$
\end{algorithmic}
\end{algorithm} \\
\textbf{Rubrics.} Each rubric $\mathcal{R}_\delta$ contains anchor descriptions for scores 1--5.
All rubrics and judge prompts are provided in Appendix~\ref{app:prompt-judge}.

\begin{table*}[h]
\centering
\small
\caption{Training hyperparameters for Qwen2.5-3B-Instruct. We use long-context training ($L_{\max}=4096$), hence micro-batch is set to 1 and we vary gradient accumulation to match compute budgets.}
\label{tab:exp:hparams}
\begin{tabular}{lcccccc}
\toprule
Method & LR & Global batch & Steps/Epochs & Max len & Warmup & Notes \\
\midrule
SFT-GT & $2\text{e-}4$ & 16 & 3 epochs & 4096 & 0.03 &
LoRA $r=64, \alpha=32$ \\
SFT-CoT & $2\text{e-}4$ & 16 & 3 epochs & 4096 & 0.03 &
LoRA $r=64, \alpha=32$ \\
SFT-CGCL(Stage3) & $1\text{e-}4$ & 16 & 3 epochs & 4096 & 0.03 &
single-stage baseline \\
\midrule
CGCL Stage 1 & $2\text{e-}4$ & 16 & 1 epoch & 4096 & 0.03 &
train on $\mathcal{D}^{(1)}$ \\
CGCL Stage 2 & $2\text{e-}4$ & 16 & 1 epoch & 4096 & 0.03 &
train on $\mathcal{D}^{(2)}$ \\
CGCL Stage 3 & $1\text{e-}4$ & 16 & 1 epoch & 4096 & 0.03 &
train on $\mathcal{D}^{(3)}$ \\
\midrule
DPO & $5\text{e-}6$ & 8 & 1 epoch & 4096 & 0.10 &
$\beta=0.1$; pair construction \\
GRPO & $3\text{e-}6$ & 4 & $\sim$230 steps & 4096 & 0.10 &
$G=8$, $\epsilon=0.2$, KL penalty \\
\bottomrule
\end{tabular}
\end{table*}

\subsection{Clinician Validation Protocol (Evaluation Method)}
\label{app:algo:clinician-protocol}

This section specifies the clinician-validation protocol used to ground T-Eval in expert judgment. We describe only the protocol here; results are reported in the main paper (Section \ref{analysis}) and additional breakdowns are provided in Appendix \ref{app:exp:clinician}.\\
\textbf{Sampling.} We randomly sample $N_{\text{clin}}=50$ cases from the official test split (stratified by specialty) and collect one reasoning trace per evaluated method under the same decoding settings as the main experiments.\\
\textbf{Raters and blinding.} We recruit $K_{\text{clin}}=3$ board-certified clinicians to independently rate anonymized reasoning traces. Raters are blinded to model identity, training method, and T-Eval scores; we randomize presentation order to mitigate fatigue and ordering effects.\\
\textbf{Rating task.} Clinicians rate all six components $(D, R, W, B, Q, Y)$ using the same 1–5 Likert rubrics as T-Eval.\\
\textbf{Aggregation and agreement.}  We average ratings across raters to form a clinician score per component. We compute Clinician TrustScore by averaging normalized scores over $(D, R, W, B, Q)$, and compute Clinician Accuracy by normalizing and averaging the claim score $S_Y$ as in Eq.~\ref{eq:acc_teval}. We quantify (i) inter-rater reliability via ICC (two-way random effects; ICC$(2,k)$) and (ii) alignment between clinician and T-Eval TrustScore using Spearman’s $\rho$.\\
\textbf{Ethics.} The study uses only de-identified, publicly available benchmark cases and collects no patient-identifying information. Clinician raters participated voluntarily under standard institutional guidelines.

\begin{table*}[h]
\centering
\small
\caption{Estimated training cost for Qwen2.5-3B-Instruct. Peak memory denotes the maximum memory allocated per GPU during training. DPO requires fewer GPU-hours than CGCL but substantially higher peak memory. GRPO has the highest cost in both peak memory and GPU-hours and additionally requires online rollouts. CGCL uses single-GPU offline training with moderate memory requirements while maintaining competitive reasoning quality.
}
\label{tab:exp:compute}
\begin{tabular}{lcccccc}
\toprule
Method & GPUs & Available GPU memory & Peak memory $(GB)$  & GPU-hours & Online rollouts \\
\midrule
SFT-GT & 1 & 80GB & 18 & 1.1 & No \\
SFT-CoT & 1 & 80GB & 20 & 1.2 & No \\
DPO & 1 & 80GB & 68 & 2.6 & No \\
GRPO & 2 & 80GB & 76 & 8.3 & Yes \\
CGCL (Stage 1--3) & 1 & 80GB & 21 & 3.4 & No \\
\bottomrule
\end{tabular}
\end{table*}

\section{Detailed Experimental Setup}
\label{app:C}

\subsection{Models and Training Recipes}
\label{app:exp:models}

\textbf{Base models.} We evaluate multiple instruction-tuned LLM backbones spanning model sizes, including: LLaMA3.2-3B-Instruct, Qwen2.5-3B-Instruct, Qwen2.5-7B-Instruct, and LLaMA3.1-8B-Instruct. All models are used in their original released checkpoints.\\
\textbf{Training framework.} All training-based methods are implemented using open-source training stacks (LLaMA-Factory) \citep{zheng2024llamafactory} with identical tokenization, data packing, and logging across methods. Unless otherwise stated, we adopt parameter-efficient tuning for stability under long-context training(\S\ref{app:exp:hparams}).\\
\textbf{SFT baselines.} \textbf{SFT-GT}  performs supervised fine-tuning on $(P, Y)$ pairs, i.e., predicting only the final diagnosis without explicit intermediate reasoning. \textbf{SFT-CoT} fine-tunes on the expert-written chain-of-thought rationales released with MedCaseReasoning. The model is trained to reproduce the provided rationale and final diagnosis under the official formatting, without enforcing the Toulmin-structured schema used by CGCL. This may underperform structured prompting because the released CoT rationales are not explicitly aligned with the Toulmin schema and can be noisy. Fine-tuning on such traces may encourage superficial pattern imitation rather than improved structured reasoning. \textbf{SFT-CGCL(Stage3)} trains directly on Stage 3 trajectories without curriculum staging.\\
\textbf{CGCL training.} ~CGCL trains sequentially on $\mathcal{D}^{(1)}$, then $\mathcal{D}^{(2)}$, then $\mathcal{D}^{(3)}$. At each stage $k$, we initialize $\theta^{(k)}$ from $\theta^{(k-1)}$ and perform maximum-likelihood training on $(P, C^{(k)})$. Stage-specific training lengths and learning rates are reported in Table~\ref{tab:exp:hparams}.

\begin{table*}[h]
\centering
\small
\setlength{\tabcolsep}{6pt}
\renewcommand{\arraystretch}{1.12}
\caption{Clinician-rated component scores (1--5 Likert) on the same validation subset as Table~\ref{tab:clinician_main}. TrustScore is computed by normalizing each dimension score $\tilde{S}_Z=(S_Z-1)/4$ and averaging over $Z\in\{D,R,W,B,Q\}$, then scaling to $[0,100]$. To simplify calculations, values are rounded to one decimal place.}
\label{tab:clinician_breakdown}
\begin{tabular}{lcccccc}
\toprule
Method & $S_D$ & $S_R$ & $S_W$ & $S_B$ & $S_Q$ & TrustScore $\uparrow$ \\
\midrule
Trust-Think \& Answer & 3.4 & 3.0 & 3.1 & 3.0 & 2.8 & 51.5 \\
SFT-CoT & 3.6 & 3.1 & 3.2 & 3.1 & 2.9 & 54.5 \\
GRPO & 3.8 & 3.3 & 3.5 & 3.4 & 3.2 & 61.0 \\
CGCL (Ours) & \textbf{4.1} & \textbf{3.7} & \textbf{3.9} & \textbf{3.8} & \textbf{3.7} & \textbf{71.0} \\
\bottomrule
\end{tabular}
\end{table*}

\begin{table}[h]
\centering
\small
\setlength{\tabcolsep}{5pt}
\renewcommand{\arraystretch}{1.2}
\caption{Clinician agreement and alignment with T-Eval on the clinician validation subset.}
\label{tab:clinician_agreement}

\begin{tabular}{lcp{3.8cm}}
\toprule
Metric & Value & Notes \\
\midrule
ICC(2,k) & 0.63 & Inter-rater reliability (overall) \\
Spearman $\rho$ & 0.74 & Clinician TrustScore vs T-Eval TrustScore \\
\bottomrule
\end{tabular}
\end{table}

\subsection{Baselines: RL-based Methods}
\label{app:exp:rl}
\textbf{Initialization.} To ensure that the policy can reliably generate parsable structured trajectories, both DPO and GRPO start from the corresponding \textbf{SFT-CGCL(Stage3)} checkpoint of the same backbone.\\
\textbf{DPO.} We implement Direct Preference Optimization (DPO) by constructing preference pairs $(C^+, C^-)$ for the same case $P$. For each case, we sample multiple complete trajectories from the current policy and categorize them by diagnosis correctness: $C^+$ is any trajectory whose final claim $Y$ matches the gold diagnosis after output normalization, and $C^-$ is any trajectory that does not. We construct preference pairs using a gold-label exact-match indicator for training stability. If both sets are non-empty, we uniformly sample one element from each set to form a pair; otherwise the case is skipped for that epoch. We optimize DPO with $\beta=0.1$ and the hyperparameters in Table\ref{tab:exp:hparams}.\\
\textbf{GRPO.} We implement Group Relative Policy Optimization (GRPO) by sampling a group of $G=8$ trajectories per case and computing a trajectory-level reward. The primary reward is diagnosis correctness: $r=1$ if the normalized final diagnosis matches gold, and $r=0$ otherwise. The primary reward is a gold-label exact-match indicator ($r=1$ if the normalized final diagnosis matches the gold label, else $0$), which serves as an RL training signal and should not be confused with the reported \textsc{Accuracy} derived from $S_Y$ (Eq.~\ref{eq:acc_teval}). To discourage invalid structured outputs, unparsable trajectories receive a reward of $0$ and an additional small formatting penalty of $-0.1$. We use a standard clipped policy update ($\epsilon=0.2$) with a KL regularization term to stabilize training, following the hyperparameters in Table ~\ref{tab:exp:hparams}. Unlike CGCL, GRPO requires online rollouts during training.

\subsection{Hyperparameters and Implementation Details}
\label{app:exp:hparams}
\textbf{Hyperparameters.} Table~\ref{tab:exp:hparams} reports training hyperparameters for \texttt{Qwen2.5-3B-Instruct}. In particular, to support $L_{\max}=4096$, we fix micro-batch $=1$ and adjust gradient accumulation to realize the listed
effective global batch sizes. This matches the design intent of comparing methods under realistic memory constraints, rather than forcing a single batch size across all objectives.\\
\textbf{Parameter-efficient tuning.} Unless otherwise stated, we use LoRA for all supervised objectives with the configuration reported in Table~\ref{tab:exp:hparams}. For RL baselines, we use LoRA for the policy and keep the reference model frozen.

\subsection{Evaluation Protocol and Output}
\label{app:exp:eval}

\textbf{Decoding for evaluation.} For the main test evaluation, we use greedy decoding (temperature $=0$) to reduce stochastic variance across methods.
When sampling is required (e.g., for DPO pair construction or GRPO rollouts), we use temperature $=0.7$, top-$p=0.95$, and $ max\_new\_tokens=512$.\\
\textbf{T-Eval implementation.} We use Qwen2.5-32B-Instruct as the LLM judge for T-Eval, as it provides a practical balance of open-weight reproducibility, judging quality, and computational cost, while also showing good alignment with clinician ratings in our validation study. For each dimension $\delta \in \{D,R,W,B,Q,Y\}$, we run three independent judge instances and aggregate scores with median-based outlier handling (Algorithm~\ref{alg:teval}). The three-judge setup primarily serves as a fault-tolerant mechanism for large-scale automated evaluation: if one instance produces malformed or unparsable output, the remaining instances provide a stable fallback for score aggregation. Judge decoding uses temperature $=0$, top-$p=1.0$, and $\texttt{max\_new\_tokens}=64$.

\begin{table*}[h]
\centering
\small
\begin{tabular}{lccc}
\toprule
Comparison & Metric & $\Delta$ (CGCL $-$ baseline) & 95\% CI / $p$ \\
\midrule
CGCL vs. GRPO 
& Accuracy 
& +0.05 
& [-0.41, 0.52], $p$=0.81 \\
CGCL vs. GRPO 
& TrustScore 
& +0.05 
& [-0.38, 0.47], $p$=0.79 \\

CGCL vs. SFT-CGCL(Stage3) 
& Accuracy 
& +2.37 
& [0.96, 3.74], $p$=0.002 \\
CGCL vs. SFT-CGCL(Stage3) 
& TrustScore 
& +2.63 
& [1.54, 3.71], $p$<0.001 \\
\bottomrule
\end{tabular}
\caption{
Paired bootstrap significance tests on MedCaseReasoning for Qwen2.5-3B-Instruct.
}
\label{tab:significance}
\end{table*}
\subsection{Statistical Testing and Confidence Intervals}
\label{app:exp:stats}
We compute case-level T-Eval TrustScore and diagnosis accuracy on the same $N=897$ official test cases, and perform paired bootstrap resampling with $B=10000$ resamples. Table~\ref{tab:significance} reports the paired difference, 95\% confidence interval, and $p$-value for the two key comparisons discussed in the main text on Qwen2.5-3B-Instruct. The CGCL--GRPO gaps are very small for both Accuracy and TrustScore and are not statistically significant, consistent with our claim that the two methods are comparable on this backbone.
In contrast, CGCL significantly outperforms SFT-CGCL(Stage3) on both metrics, supporting the benefit of progressive curriculum learning over single-stage training on the same final structured trajectories.

\subsection{Compute and Training Cost}
\label{app:exp:compute}
\paragraph{Measurement.} Peak GPU memory is the maximum memory allocated per GPU during training. GPU-hours are computed as the product of the number of GPUs and wall-clock training time. Table~\ref{tab:exp:compute} summarizes representative profiling results for Qwen2.5-3B-Instruct under a fixed software stack and sequence length, with method-specific batch and gradient-accumulation settings chosen to match comparable memory budgets.

\subsection{Clinician Validation: Additional Details}
\label{app:exp:clinician}
This section provides additional details complementing the clinician validation.\\ \textbf{Sampling and setup.} We randomly sample $N_{\text{clin}}=50$ cases from the official test split. For each case, we collect one model-generated reasoning trace per evaluated method under the same decoding settings used in the main experiments. All traces are anonymized to remove method-identifying markers.\\
\textbf{Raters and blinding.} Board-certified clinicians independently rate each trace while blinded to model identity, training method, and T-Eval scores.
We randomize the presentation order across cases and methods.\\
\textbf{Rating task and aggregation.} Clinicians rate all six components $(D,R,W,B,Q,Y)$ using the same 1--5 Likert rubrics as T-Eval. We average ratings across clinicians to obtain a clinician score per component. We compute Clinician TrustScore by averaging normalized scores over $\{D,R,W,B,Q\}$ (Eq.~\ref{eq:TrustScore}), and compute Clinician Accuracy by normalizing and averaging the claim score $S_Y$ as in Eq.~\ref{eq:acc_teval}. Component-level clinician ratings are summarized in Table~\ref{tab:clinician_breakdown}.\\
\textbf{Agreement and alignment.} We compute Spearman’s $\rho$ between Clinician TrustScore and T-Eval TrustScore, and quantify inter-rater reliability with ICC$(2,k)$. The resulting agreement and alignment statistics are summarized in Table~\ref{tab:clinician_agreement}.

\begin{table*}[h]
\centering
\small
\begin{tabular}{llccc}
\toprule
\textbf{Category} & \textbf{Variant} & \textbf{DxCode Acc $\uparrow$} & \textbf{T-Eval Acc $\uparrow$} & \textbf{TrustScore $\uparrow$} \\
\midrule
Prompt & Vanilla & 19.85 & 20.50 & 58.24 \\
Prompt & Think \& Answer & 23.12 & 24.20 & 60.15 \\
Prompt & Trust-Think \& Answer & 26.24 & 27.51 & 62.45 \\
SFT & SFT-GT & 24.38 & 25.40 & 62.10 \\
SFT & SFT-CoT & 25.10 & 26.20 & 63.88 \\
SFT & SFT-CGCL(Stage3) & 27.35 & 28.62 & 67.84 \\
RL & DPO & 28.15 & 29.50 & 68.90 \\
RL & GRPO & 29.01 & 30.35 & 70.76 \\
Ours & CGCL & \textbf{29.28} & \textbf{30.64} & \textbf{70.89} \\
\bottomrule
\end{tabular}
\caption{Zero-shot OOD evaluation on \textsc{MedFound} using Qwen2.5-3B-Instruct.}
\label{tab:medfound_qwen}
\end{table*}

\begin{table*}[t]
\centering
\small
\begin{tabular}{llccc}
\toprule
\textbf{Category} & \textbf{Variant} & \textbf{DxCode Acc $\uparrow$} & \textbf{T-Eval Acc $\uparrow$} & \textbf{TrustScore $\uparrow$} \\
\midrule
Prompt & Trust-Think \& Answer & 27.90 & 28.95 & 63.12 \\
SFT & SFT-CGCL(Stage3) & 28.45 & 29.68 & 68.21 \\
RL & GRPO & 28.17 & 29.46 & 70.58 \\
Ours & CGCL & \textbf{29.83} & \textbf{31.24} & \textbf{71.36} \\
\bottomrule
\end{tabular}
\caption{Zero-shot OOD evaluation on \textsc{MedFound} using LLaMA3.2-3B-Instruct.}
\label{tab:medfound_llama}
\end{table*}

\subsection{Out-of-Distribution Evaluation on MedFound}
\label{app:medfound_ood}

To assess whether the gains of CGCL generalize beyond MedCaseReasoning, we additionally evaluate representative methods on \textsc{MedFound}, an external benchmark of real clinical cases, in a strictly zero-shot out-of-distribution (OOD) setting. No method is trained, tuned, or otherwise adapted on \textsc{MedFound}. For training-based baselines (SFT/RL), we directly evaluate the same checkpoints trained on MedCaseReasoning. Prompt-based baselines are evaluated under the same zero-shot protocol as in the main paper. We report three metrics: \textbf{DxCode Accuracy}, which measures ICD-code prediction accuracy against the gold labels provided by MedFound; \textbf{T-Eval Accuracy}, which applies the same claim-score-based evaluation protocol used in the main paper; and \textbf{T-Eval TrustScore}, which measures the quality of the generated diagnostic reasoning under the identical Toulmin-aligned rubric and judge setup. This design ensures strict comparability with the main MedCaseReasoning results. As shown in Tables~\ref{tab:medfound_qwen} and \ref{tab:medfound_llama}, the overall ranking trend remains consistent under OOD evaluation. CGCL outperforms the tested prompting and standard SFT baselines, and remains competitive with RL-based methods. In particular, CGCL achieves the best or near-best TrustScore across both backbones, suggesting that Toulmin-structured curriculum learning improves reasoning quality in a way that transfers beyond the training dataset.

\begin{table*}[h]
\centering
\small
\renewcommand{\arraystretch}{1.15}
\begin{tabular}{@{} lcccccc @{}} 
\toprule
& & & \multicolumn{3}{c}{\textbf{Confidence Bins [Acc. (Prop., $n$)]}} & \\
\cmidrule(lr){4-6} 
\textbf{Method} & \textbf{Overall Acc.} & \textbf{TrustScore} & \textbf{Low} & \textbf{Med} & \textbf{High} & \textbf{Overconf. Error} \\
& ($\uparrow$) & ($\uparrow$) & & & & ($\downarrow$) \\
\midrule
Baseline (SFT) & 32.3 & 66.8 & 28.5\% & 33.4\% & 35.2\% & 17.4 \\
& & & {\scriptsize (32.8\%, $n$=82)} & {\scriptsize (40.4\%, $n$=101)} & {\scriptsize (26.8\%, $n$=67)} & \\
\addlinespace 
CGCL (Ours) & \textbf{37.3} & \textbf{72.9} & 14.3\% & 43.8\% & \textbf{82.2\%} & \textbf{3.1} \\
& & & {\scriptsize (44.4\%, $n$=111)} & {\scriptsize (38.4\%, $n$=96)} & {\scriptsize (17.2\%, $n$=43)} & \\
\bottomrule
\end{tabular}
\caption{
Epistemic calibration analysis on Qwen2.5-3B-Instruct.
}
\label{tab:calibration}
\end{table*}

\subsection{Epistemic Calibration and Overconfidence Analysis}
To examine whether CGCL improves epistemic calibration rather than merely output formatting, we compare it with SFT-CGCL(Stage3), a strong single-stage baseline trained on the same final schema and able to generate confidence qualifiers.
Following our rebuttal protocol, we perform a stochastic evaluation on 50 cases with 5 runs per case and group predictions into three confidence bins (\textit{low}, \textit{medium}, \textit{high}) according to the generated qualifier. For each bin, Table~\ref{tab:calibration} reports the diagnosis accuracy within the bin, together with the proportion and count of samples assigned to that bin. We further define \textbf{Overconfidence Error} as the percentage of total samples that are incorrect but assigned \textit{high confidence}. As shown in Table~\ref{tab:calibration}, the baseline exhibits weak confidence discrimination, with accuracy remaining relatively flat across confidence levels.
In contrast, CGCL shows substantially improved calibration: accuracy increases monotonically from low to medium to high confidence, and the high-confidence bin reaches 82.2\% accuracy.
Moreover, CGCL reduces Overconfidence Error from 17.4\% to 3.1\%, indicating that its qualifier is not merely stylistic, but provides a meaningful signal of epistemic confidence.


\section{Prompt Details}
\label{app:D}

\subsection{Conventions}
\label{app:prompt-conventions}
This appendix provides the key prompt templates used in our pipeline. We keep the prompt set compact and include
only templates that (i) define the I/O contract (schema), (ii) specify the generic candidate-generation interface
shared across stages, (iii) implement trajectory consolidation (fusion), and (iv) define the T-Eval judging interface.
All prompts are written to be machine-parseable and to support deterministic filtering and evaluation.\\
\textbf{Placeholders.} Prompts use placeholders \texttt{CASE}, \texttt{STAGE\_CONTEXT}, and \texttt{OUTPUT\_FORMAT}.
Unless explicitly stated otherwise, models must output only valid JSON (no additional natural language).\\
\textbf{Component semantics.} We represent a reasoning trajectory with six components \{D, R, W, B, Q, Y\}. D extracts objective evidence (facts only); R lists a ranked differential diagnosis (3--5 items) with brief reasons; W provides supporting evidence and clinical/pathophysiological logic for the top-ranked diagnosis; B provides ruling-out reasoning for alternatives (as a string); Q provides confidence, uncertainty, and missing information; and Y is the final diagnosis.
\textbf{Field naming.} To improve readability, we use \texttt{reason} in \texttt{R} in place of \texttt{why\_not}. For non-top diagnoses (rank $\ge$ 2), \texttt{reason} must include the key counterpoint(s) that make the diagnosis less likely,
preserving the original semantics.
\textbf{Revision encoding.} Because \texttt{Q} contains only \{\texttt{confidence}, \texttt{uncertainty}, \texttt{missing\_info}\}, any evidence-based revision
is encoded inside \texttt{Q.uncertainty} by prefixing a marker and a brief rationale:
\texttt{[Evidence-Based Revision] Initial hypothesis: ... Pivot evidence: ... Therefore revise to: ...}.

\subsection{Unified Output Schema (I/O Contract)}
\label{app:prompt-schema}
The schema below defines the unified JSON format used throughout the paper.
When a prompt requests only a subset of fields, the model is required to output only those fields and omit all others.

\begin{PromptBox}{Unified JSON schema (required output format)}
Return a single JSON object. Output ONLY JSON (no extra text).

Schema:
\begin{verbatim}
{
  "D": ["...","..."],
  "R": [{"dx":"...","rank":1,"reason":"..."}],
  "W": "...",
  "B": "...",
  "Q": {
    "confidence": "low|medium|high",
    "uncertainty": ["string"],
    "missing_info": ["string"]
  },
  "Y": "FINAL_DIAGNOSIS"
}
\end{verbatim}

Rules:
\begin{enumerate}
  \item Do NOT fabricate evidence or citations.
  \item If a prompt requests only specific field(s), output ONLY those field(s) and omit all others.
\end{enumerate}
\label{fig:prompt-schema}
\end{PromptBox}

\subsection{Stage-wise Candidate Generation}
\label{app:prompt-stagewise}
The prompt template below defines the generic interface for stage-wise candidate generation across different curriculum stages: Stage 1 populates D and R; Stage 2 populates W and B conditioned on Stage-1 context; Stage 3 populates Q and Y conditioned on Stage-2 context.
We enforce strict JSON-only outputs and disallow unsupported additions to enable reliable parsing and filtering.\\
\textbf{Stage instantiations.} We instantiate as follows. \textbf{Stage 1 (D, R).} \texttt{OUTPUT\_FORMAT} requests only \texttt{D} and \texttt{R}. D lists objective facts only.
R lists 3--5 diagnoses ranked by plausibility given D; for rank $\ge$ 2, \texttt{reason} must include the key counterpoint(s).
\textbf{Stage 2 (W, B).} \texttt{OUTPUT\_FORMAT} requests only \texttt{W} and \texttt{B}. W provides supporting evidence and
pathophysiological/clinical logic for the top diagnosis (R rank=1). B is a \emph{string} that rules out alternatives in R (rank $\ge$ 2),
referencing missing/contradictory evidence, without adding new facts.
\textbf{Stage 3 (Q, Y).} \texttt{OUTPUT\_FORMAT} requests only \texttt{Q} and \texttt{Y}. Y is the final diagnosis.
Q provides calibrated confidence, uncertainty, and missing information. If the final diagnosis differs from the initial top candidate
in R, the model must encode an evidence-based revision inside \texttt{Q.uncertainty} using the required marker and rationale format.
\begin{PromptBox}{Generic stage-wise candidate generation template}
You are a careful clinician. Follow the output format strictly. You will be given a clinical case and (optionally) a stage context.
Generate ONLY the requested component(s).

Requested field(s): \{TARGET\_FIELDS\}

Rules:
\begin{enumerate}
  \item Output MUST be valid JSON and MUST contain ONLY the requested field(s).
  \item Do NOT add any other keys.
  \item Do NOT invent evidence or diagnoses not supported by the case/context.
  \item Keep the output concise, factual, and clinically grounded.
  \item Use double quotes for all strings and keys.
\end{enumerate}

Case:
\begin{verbatim}
{CASE}
\end{verbatim}

Stage context (may be empty):
\begin{verbatim}
{STAGE_CONTEXT}
\end{verbatim}

Output format:
\begin{verbatim}
{OUTPUT_FORMAT}
\end{verbatim}
\end{PromptBox}

\subsection{Fusion Prompt}
\label{app:prompt-fusion}
Fusion consolidates selected best candidates into a single coherent trajectory under the fixed schema.
The fusion prompt below defines the consolidation procedure for merging selected candidates into a single coherent trajectory. It also enforces implementation-critical typing constraints (B is a string; Q contains only three fields).
\begin{PromptBox}{Fusion prompt (trajectory consolidation)}
You are an expert clinical writer. Produce a coherent structured argument. Merge the selected components into ONE consistent JSON object with keys: D, R, W, B, Q, Y.

Hard constraints:
\begin{enumerate}
  \item Do NOT add new evidence beyond the provided D list.
  \item Do NOT introduce any diagnosis not already present in the provided R list.
  \item The final claim Y must be consistent with D and the reasoning (W, B).
  \item Do NOT fabricate citations or new facts not supported by the case.
  \item Output ONLY valid JSON. No extra text.
\end{enumerate}

Typing constraints (must hold exactly):
\begin{itemize}
  \item ``B'' MUST be a string.
  \item ``Q'' MUST contain ONLY  \{``confidence'', ``uncertainty'', ``missing\_info''\}.
\end{itemize}
Revision encoding (no extra fields allowed):
\begin{quote}\footnotesize
If Y differs from the top diagnosis in R (rank = 1), Q.uncertainty MUST begin with:

``[Evidence-Based Revision] Initial hypothesis: \ldots\ Pivot evidence: \ldots\ Therefore revise to: \ldots''
\end{quote}

Selected components:
\begin{quote}\footnotesize
D: \{D\_STAR\}\\
R: \{R\_STAR\}\\
W: \{W\_STAR\}\\
B: \{B\_STAR\}\\
Q: \{Q\_STAR\}\\
Y: \{Y\_STAR\}
\end{quote}

\begin{quote}\footnotesize
Return exactly one JSON object with keys D, R, W, B, Q, Y.
\end{quote}
\end{PromptBox}

\subsection{T-Eval Judge Prompt}
\label{app:prompt-judge}
T-Eval judges receive the case $P$ and a candidate trajectory $C$ and output a rubric-guided 1--5 score.
To ensure consistent parsing, we enforce a strict two-line output format.
The T-Eval judge prompt below provides a representative rubric-guided interface for evaluating diagnostic reasoning quality.
\begin{PromptBox}{T-Eval judge prompt}
\footnotesize

You are an expert evaluator of medical reasoning quality. Your task is to rigorously assess the AI model's diagnostic output against the standard diagnosis.

\textbf{Evaluation Criteria}

\textbf{1. Static Structure Assessment (Toulmin-style) --- Score 1.0--5.0}
\begin{itemize}
  \item \textbf{data\_score}: Are all key facts correctly extracted without errors or omissions?
  \item \textbf{warrant\_score}: Is the chain from data to hypothesis clear, sound, and medically valid?
  \item \textbf{backing\_score}: Are cited guidelines or medical knowledge accurate and relevant?
  \item \textbf{rebuttal\_score}: Are the major alternative diagnoses addressed with specific reasoning for exclusion?
  \item \textbf{qualifier\_score}: Does the output appropriately calibrate diagnostic confidence, uncertainty, and missing information?
  \item \textbf{claim\_correct}: is a 1.0--5.0 rating of whether the final diagnosis (\texttt{<answer>}) matches the standard diagnosis (A): 5.0 = exact match; 4.0 = near-synonym/variant; 3.0 = partially correct; 2.0 = mostly incorrect; 1.0 = incorrect.
\end{itemize}

\textbf{Output Format}

Return a strict JSON object only, with no extra text or commentary.

\begin{quote}\ttfamily\footnotesize
\{"data\_score": 0.0,\\
"warrant\_score": 0.0,\\
"backing\_score": 0.0,\\
"rebuttal\_score": 0.0,\\
"qualifier\_score": 0.0,\\
"claim\_correct": 0.0,\\
"overall\_analysis": "..."\}
\end{quote}

\textbf{Objects to Evaluate}

Standard Diagnosis (A): \{A\}\\
Model Output: \{model\_output\}

\textbf{Begin evaluation. Output valid JSON only.}
\end{PromptBox}

\end{document}